# ADAPTIVE NOISE REDUCTION SCHEME FOR SALT AND PEPPER

Tina Gebreyohannes[1] and Dong-Yoon Kim[2]

[1]Department of Computer Engineering, Ajou University, Suwon, South Korea
tinosaturn@gmail.com
[2]Departent of Computer Engineering, Ajou University, Suwon, South Korea
dykim@ajou.ac.kr

## ABSTRACT

*In this paper, a new adaptive noise reduction scheme for images corrupted by impulse noise is presented. The proposed scheme efficiently identifies and reduces salt and pepper noise. MAG (Mean Absolute Gradient) is used to identify pixels which are most likely corrupted by salt and pepper noise that are candidates for further median based noise reduction processing. Directional filtering is then applied after noise reduction to achieve a good tradeoff between detail preservation and noise removal. The proposed scheme can remove salt and pepper noise with noise density as high as 90% and produce better result in terms of qualitative and quantitative measures of images.*

## 1. INTRODUCTION

The two most common types of noise in image processing are Gaussian noise and Impulse noise, also known as salt and pepper noise. This type of noise may appear in digital image due contaminated impulse noise, which is caused by malfunctioning pixels in camera sensors, faulty memory location in hardware, or transmission in noisy channel [9]. Salt and pepper noise scattered throughout the image in such a way pixels can take only the maximum and minimum values (0 and 255 respectively) in the dynamic range.

Many researches have been conducted and numerous algorithms were proposed to remove salt and pepper noise. Among these noise reduction techniques, majority splits the noise removal procedures into preliminarily detection of pixels corrupted by impulse noise followed by filtering the noise detected on the previous phase [8]. Standard median filter (SMF) [7] was one of famous among others due to its great denoising performance and computational efficiency. But since the conventional median filter applies the median operation to each pixel whether it is corrupted or not, it suffers from preserving some details of the image as the noise density increases. More improved algorithms such as adaptive median filter (AMF) [2], decision-based algorithm (DBA) [1] and convolution-based algorithm (CBA) [3] mainly focus on noise detector. Pixels detected by the noise detector will be considered as noise and shall further be processed in their respective noise reduction scheme. Having such mechanism with phases would highly preserve the details of the image and save the restored image from having blurred and distorted feature.

The proposed algorithm in this paper also greatly focuses on how to effectively detect the salt and pepper noise and efficiently restore the image. The mechanism adopted by the proposed scheme consists of three phases. The first phase determines whether a pixel is noise or not based on some predefined threshold and calculated values. Once pixels are detected as noise in previous phase, their new value will be estimated and set in noise reduction phase. Finally a conditional image enhancement phase will be conducted for those images which have been corrupted with high density noise to preserve edges and details of the restored image. This makes the proposed algorithm to have an outstanding performance even at noise density as high as 90%.

The rest of this paper is organized as follows. Section 2 describes the proposed scheme. Section 3 presents the experimental results. And the last section, Section 4 concludes this paper.

## 2. PROPOSED SCHEME

In this paper, three steps adaptive noise reduction scheme is proposed for salt and pepper noise.

For $(i,j) \in X = \{1,\ldots,N\} \times \{1,\ldots,N\}$, let $x_{ij}$ be the gray level of the original image X. The observed grey level at pixel location $(i,j)$ is given by :

$$f(x_{ij}) = \begin{cases} 0 & \text{with } p/2 \\ x_{ij} & \text{with } 1-p \\ 255 & \text{with } p/2 \end{cases} \quad (1)$$

Where p is noise ratio contaminated in the image.

### 2.1. Noise detection scheme

To classify corrupted and uncorrupted pixels Mean Absolute Gradient (MAG) will be applied. A small MAG indicates a flat region, and a large MAG usually indicates complex region or impulse noise region. MAG is defined as follows:

$$MAG = \frac{1}{N-1}\sum_{i=0}^{N-1} |F(0) - F(i)| \quad (2)$$

Where F(i) denotes the intensity value of pixel in a region, F(0) is the intensity of the pixel in the center as shown in Fig. 1 and N is the number of pixels in the region.

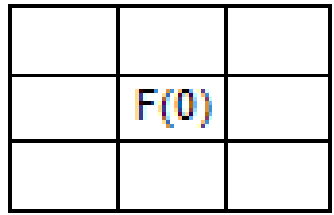

Figure 1. 3x3 MAG window

$$z_{ij} = \begin{cases} 1 & MAG > T \\ 1 & x_{ij} = 0 \\ 1 & x_{ij} = 255 \\ 0 & MAG \leq T \end{cases} \quad (3)$$

A reasonable threshold T can be determined using computer simulation. If $Z_{ij} = 1$, then the pixel $X_{ij}$ is marked as noise candidate; otherwise the pixel $X_{ij}$ is noise free.

## 2.2. Noise Reduction Scheme

For the noisy pixel detected, we use 3x3 window M as shown in eq. 4.

$$M = \begin{bmatrix} x_{i-1j-1} & x_{ij-1} & x_{i+1j-1} \\ x_{i-1j} & x_{ij} & x_{i+1j} \\ x_{i-1j+1} & x_{ij+1} & x_{i+1j+1} \end{bmatrix} \quad (4)$$

By sorting elements in M according to their distance from the center pixel, we get the sorted sequence as follow:

i. Sort the five element $\{x_{ij}, x_{i-1j-1}, x_{i+1j-1}, x_{i-1j+1},$ and $x_{i+1j+1}\}$ in ascending order, then we get $\{a_1, a_2, a_3, a_4, a_5\}$ where $\{a_1 \leq a_2 \leq a_3 \leq a_4 \leq a_5\}$.

   Median = $a_3$.

ii. Sort the other four elements including Median from step (i) $\{x_{ij-1}, x_{i-1j}, x_{i+1j}, x_{ij+1}$ and Median $\}$ in ascending order, then we get $\{b_1, b_2, b_3, b_4, b_5\}$ where $\{b_1 \leq b_2 \leq b_3 \leq b_4 \leq b_5\}$.

   Median = $b_3$.

Then the restored image y with $y_{ij}$ is calculated as follow:

$$y_{ij} = \begin{cases} \text{Median} & \text{if } b_2 = 0 \text{ or } b_5 = 255 \\ \text{Average}(b_4, b_5) & \text{if } b_5 = 0 \\ b_5 & \text{if } b_4 = 0 \\ \text{Average}(b_1, b_2) & \text{if } b_2 = 255 \\ b_1 & \text{if } b_5 = 255 \end{cases} \quad (5)$$

## 2.3. Image Enhancement

To preserve the details and edges of the restored image we apply directional filtering which will enhance the image quality. In order to do so, we apply standard deviation on the noisy image to determine whether directional filtering is needed to be applied or not. Standard deviation is defined as follows:

$$\sigma = \sqrt{\frac{1}{N}\sum_{i=1}^{N}(F(i) - \mu)^2} \quad (6)$$

Where F(i) denotes the intensity values of the noisy image, N is the number of pixels in the image and μ is mean of the noisy image. If standard deviation is greater than the threshold T, then directional filter will be applied to the processed image accordingly.

A pixel $X_k$ in the image is partitioned into $(Y_1, Y_2, Y_3, Y_4)$ where $\{X_1, X_2, X_4\}$ belongs to $Y_1$, which are in the upper left portion of the mask.

$$d1 = |x_1 - x_2|$$
$$d2 = |x_1 - x_4|$$
$$d3 = |x_4 - x_2|$$

So that (7)

$$y_1 = \begin{cases} avg(x_1, x_2) & \text{if } d1 = \min(d1, d2, d3) \\ avg(x_1, x_4) & \text{if } d2 = \min(d1, d2, d3) \\ avg(x_4, x_2) & \text{if } d3 = \min(d1, d2, d3) \end{cases}$$

$Y_2$, $Y_3$ and $Y_4$ with $\{X_2, X_3, X_6\}$, $\{X_4, X_7, X_8\}$ and $\{X_6, X_8, X_9\}$ respectively will also be calculated in the same manner.

$$X_K = \text{Average}(Y_1, Y_2, Y_3, Y_4) \qquad (8)$$

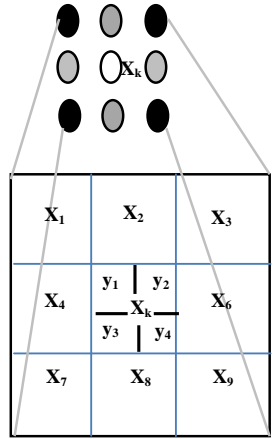

Figure 2. Directional filtering window

## 3. EXPERIMENTAL RESULTS

In this section, the experimental result of the proposed scheme is presented. A variety of simulations are carried out on 512x512 grey scale Lena image to verify the performances of various restoration methods, including AMF[2], DBA[1], CBA[3] and the proposed scheme. In the simulation, images are corrupted by "salt" (255) and "pepper" (0) noise with equal probability. The noise density is varied from 10% to 90% with 10% incremental, and the restoration performance is quantitatively measured by peak signal-to-noise ratio (PSNR). Peak signal to noise ratio (PSNR) is defined as:

$$\text{PSNR in dB} = 10 \log_{10}\left(\frac{255^2}{MSE}\right) \qquad (9)$$

where 255 is the peak signal for 8 bit image and MSE is mean square error given by:

$$MSE = \frac{1}{n^2}\left[\sum_{i=0}^{N-1}\sum_{j=0}^{N-1}(f_{ij} - f'_{ij})^2\right] \qquad (10)$$

Where $f_{ij}$ and $f'_{ij}$ are the pixel values at position (i,j) of original and processed image respectively.

The PSNR comparison of various algorithms including DBA, AMF, CBA and the proposed scheme are presented in Figure 3 graphically. Figure 4 and Figure 5 show the subjective visual qualities of the filtered image using various algorithms for the image "Lena" corrupted by 50% and 80% impulse noise respectively.

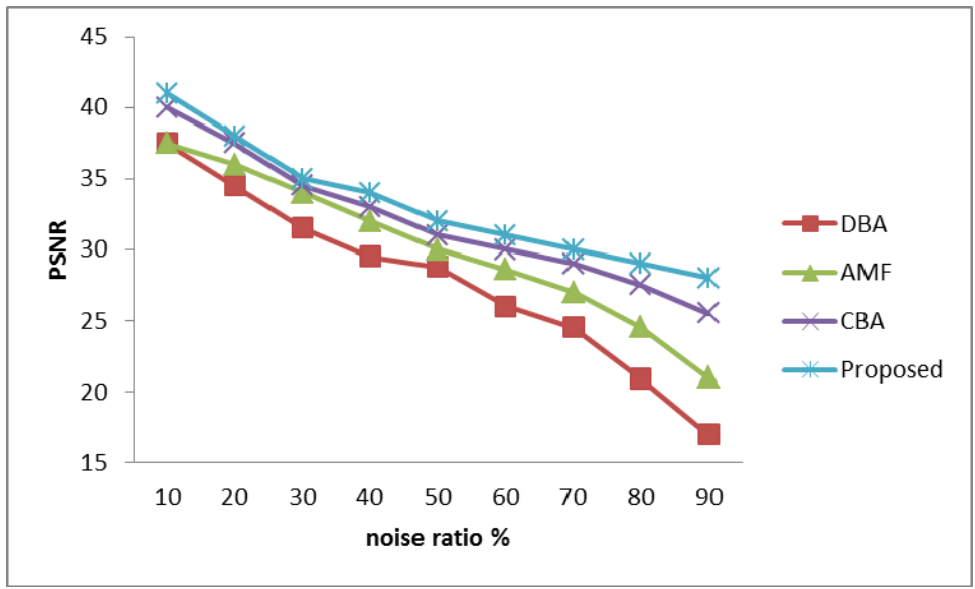

Figure 3.Comparison of various scheme in PSNR for Lena image with various ratio of salt and pepper noise

When the noise density is greater than 50%, other algorithm's PSNR drops sharply unlike the proposed algorithm which doesn't show drastic change in PSNR values. On the other hand, to further demonstrate visual quality of images for example Figure 5 presents restoration of Lena image initially corrupted with 80% impulse noise density. For image processed with DBA, from the subjective quality of the image one can observe that some of the impulse noises are not fully suppressed while blurring and distortion are serious. AMF and CBA suppress much of the noise where CBA fails to preserve edge information while maintaining details of the image unlike AMF which suffers to maintain both. Both the performance of PSNR and the subjective visual qualities of our method outperform other algorithms i.e. DBA, AMF and CBA in all cases. Specially as noise density becomes high and images are greatly corrupted by the impulse noise, our scheme shows remarkable result after images are being processed.

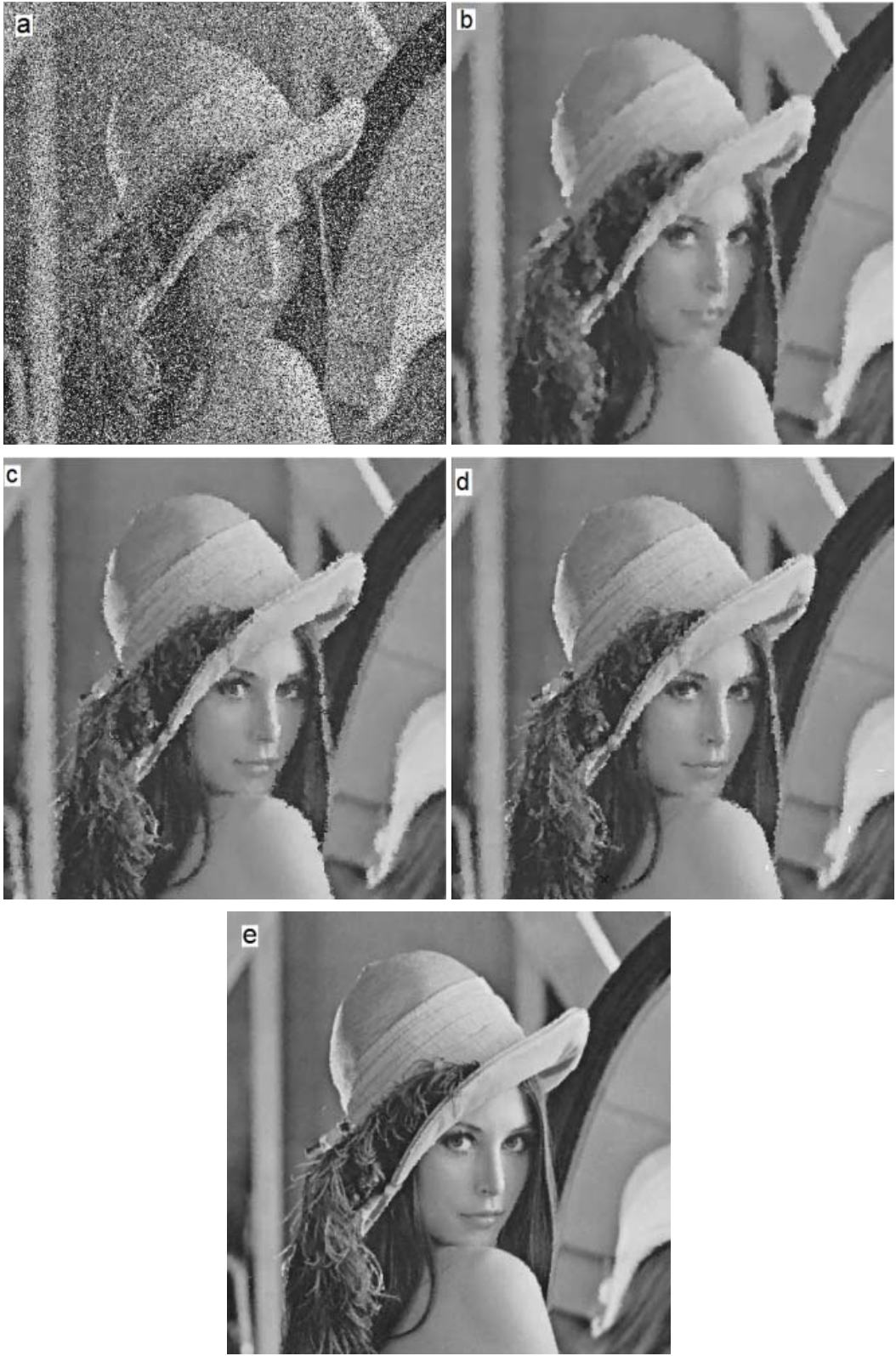

Figure 4.Image restoration results of Lena image. (a) 50% noise corrupted, (b) DBA, (c) AMF, (d) CBA, (e) proposed

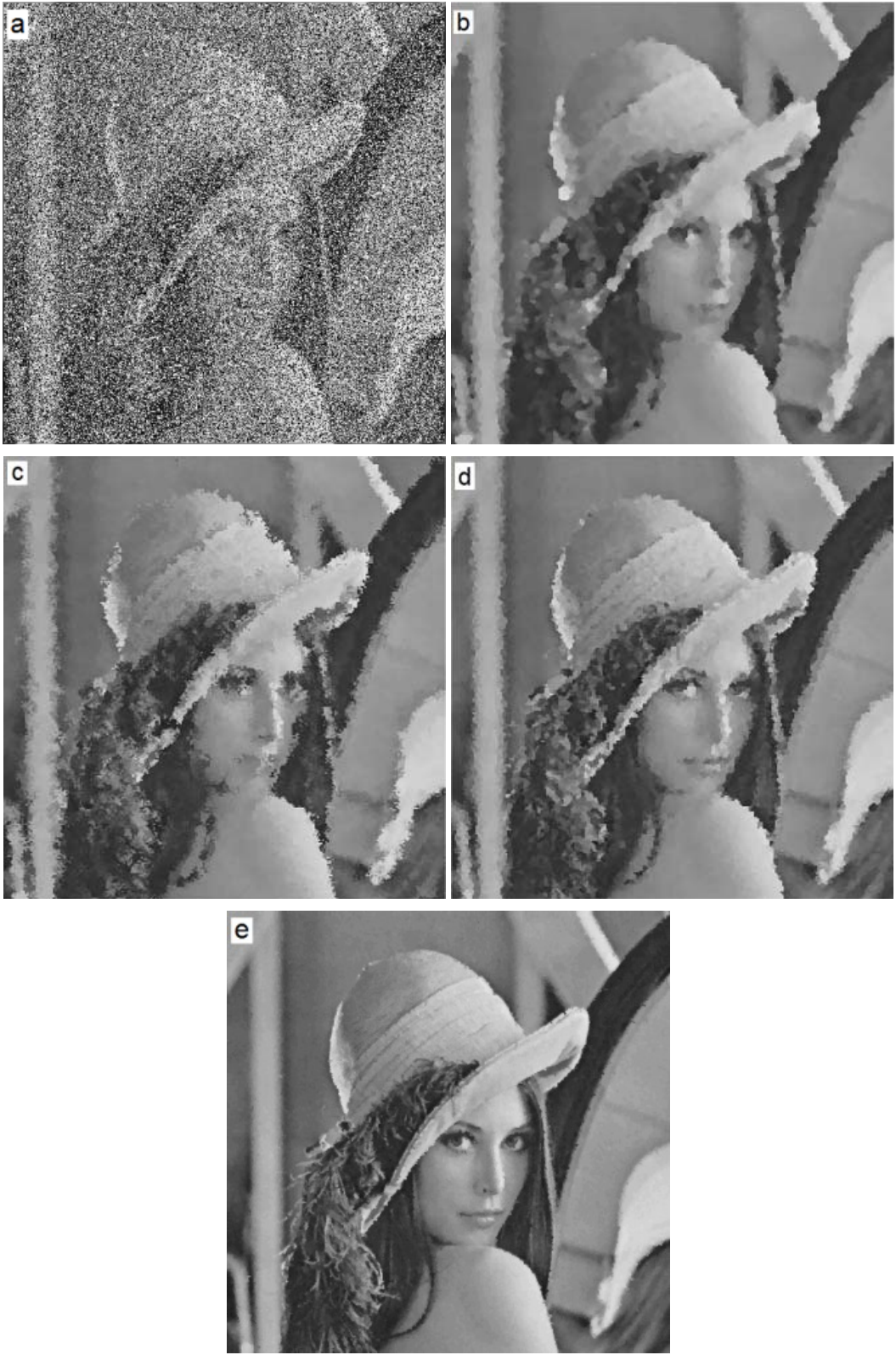

Figure 5. Image restoration results of the Lena image. (a) 80% noise corrupted, (b) DBA, (c) AMF, (d) CBA, (e) proposed

## 4. CONCLUSION

In this paper, a new adaptive noise reduction scheme for removing salt and pepper noise is proposed. The first phase of the scheme efficiently identifies impulse noise while the other is to remove the noise from the corrupted image that is followed by image enhancement scheme to preserve the details and image quality. As per the experimental results, the proposed algorithm yields good filtering result using efficient noise detection mechanism. This is observed by numerical measurements like PSNR and visual observations through the experiments conducted.

## ACKNOWLEDGEMENTS

The support of the Agency for Defence Development Korea is gratefully acknowledged.

## Authors

**Tina Gebreyohanes** received her B.Sc. degree in 2007 in Computer Science from Addis Ababa University, Ethiopia. She worked for a Telecom company called ZTE(H.K) Ethiopian branch as Fixed Line Next Generation Network(FLNGN) Engineer. She is currently M.Sc. student in Information and communication Technology Faculty, Computer Engineering Department in Ajou University. Her research interest includes image processing and data compression.

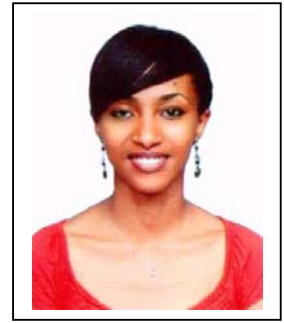

**Dong-Yoon Kim** received his B.Sc. degree in 1974 in Mathematics from Seoul National University, M.S. degree in 1976 in Computer Science from Korea Advance Institute of Science and Technology, and Ph.D., in 1985 in Applied Mathematics from Massachusetts Institute of Technology. From 1976 to 1991, he was with Agency for Defence Development Korea. In 1991 he joined Ajou University. He served as a President of Korea Institute for Information Scientists and Engineers in 2006. Also he was a Vice President of International Federation for Information Processing from 2004 to 2007. His research interests include image processing and data mining.

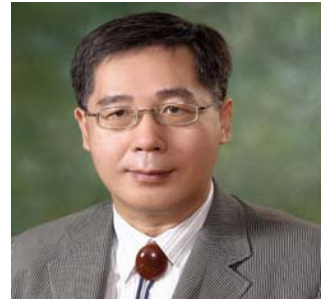